\documentclass{article}

\usepackage[preprint]{neurips_2026}

\PassOptionsToPackage{numbers, sort&compress}{natbib}
\usepackage[preprint]{neurips_2026}

\usepackage[utf8]{inputenc} %
\usepackage[T1]{fontenc}    %
\usepackage[colorlinks]{hyperref}       %
\usepackage{url}            %
\usepackage{booktabs}       %
\usepackage{amsfonts}       %
\usepackage{nicefrac}       %
\usepackage{microtype}      %
\usepackage{xcolor}         %

\usepackage{graphicx}
\usepackage{subcaption}
\usepackage{amsmath}
\usepackage{amssymb}
\usepackage{mathtools}
\usepackage{amsthm}
\usepackage[capitalize,noabbrev]{cleveref}
\usepackage{multirow} %
\usepackage{wrapfig}
\usepackage{enumitem}
\usepackage{bbm}
\usepackage{algorithm}
\usepackage[noend]{algpseudocode}
\usepackage{listings}
\usepackage{xcolor}
\usepackage{framed}
\usepackage{wrapfig}

\definecolor{myblue}{HTML}{237ab5}
\hypersetup{urlcolor=myblue,citecolor=myblue,linkcolor=myblue}

\theoremstyle{plain}

\theoremstyle{definition}

\theoremstyle{remark}

\usepackage{amsmath,amsfonts,bm,cancel}

\def\eqref#1{equation~\ref{#1}}

\def\1{\bm{1}}

\DeclareMathAlphabet{\mathsfit}{\encodingdefault}{\sfdefault}{m}{sl}
\SetMathAlphabet{\mathsfit}{bold}{\encodingdefault}{\sfdefault}{bx}{n}

\def\gB{{\mathcal{B}}}

\def\gG{{\mathcal{G}}}

\def\gO{{\mathcal{O}}}

\newcommand{\E}{\mathbb{E}}

\DeclareMathOperator*{\argmax}{arg\,max}

\definecolor{positive}{rgb}{0,0.55,0}
\definecolor{negative}{rgb}{0.75,0,0}

\theoremstyle{definition}
\newtheorem*{restatement}{Restatement}
\newenvironment{restate}[1][]
{\begin{framed}\begin{restatement}[#1]\normalfont\itshape}
{\end{restatement}\end{framed}}

\usepackage[textsize=tiny]{todonotes}

\title{Self-Supervised Goal-Reaching Results in Multi-Agent Cooperation and Exploration}

\author{%
  Chirayu Nimonkar\thanks{Equal contribution. Correspondence to: \texttt{\{chirayu, shlokshah\}@princeton.edu}.} \\
  Princeton University \\
  \And
  Shlok Shah\footnotemark[1] \\
  Princeton University \\
  \And
  Catherine Ji \\
  Princeton University \\
  \And
  Benjamin Eysenbach \\
  Princeton University \\
}

\begin{document}

\maketitle

\begin{abstract}
For groups of autonomous agents to achieve a particular goal, they must engage in coordination and long-horizon reasoning. Rather than relying on complex reward functions and explicit cooperation mechanisms, we ask what minimal ingredients are required for effective coordination and exploration to emerge in multi-agent settings. We investigate this question through self-supervised goal-reaching, where agents aim to maximize the likelihood of visiting a goal state rather than maximizing a reward. Despite a sparse feedback signal, we present empirical results that show self-supervised goal-reaching techniques enable agents to learn from such feedback. On MARL benchmarks, self-supervised goal-reaching outperforms alternative approaches that have access to the same sparse reward signal. Furthermore, we empirically demonstrate that multi-agent self-supervised goal-reaching approaches can be more robust than single-agent strategies. While there is no \emph{explicit} exploration mechanism, this approach explores nontrivial intermediate coordination strategies in sparse settings where alternative approaches fail to achieve a single success.\footnote{Project website with code and videos: 
\url{https://chirayu-n.github.io/gcmarl}}

\end{abstract}

\section{Introduction}

\begin{wrapfigure}[16]{R}{0.45\linewidth}
    \centering
    \vspace{-1.5em}
    \includegraphics[width=\linewidth,clip,trim={5cm 2cm 2cm 5cm}]{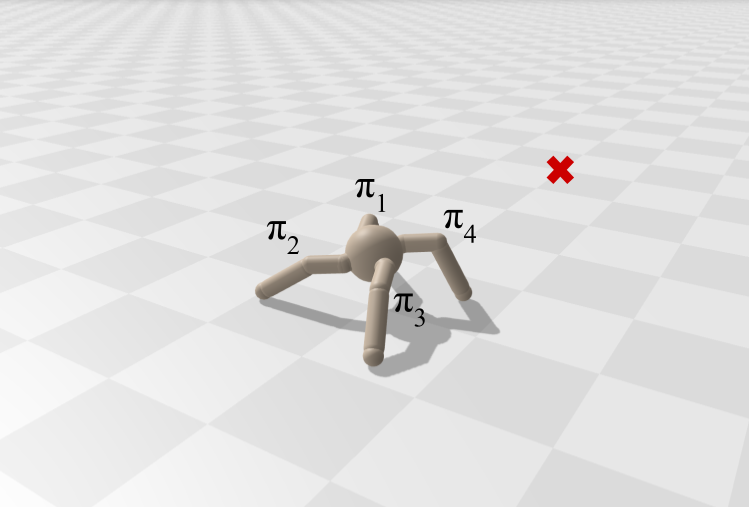}
    \vspace{-1.5em}
    \caption{\footnotesize In the \textbf{multi-agent goal-reaching} problem, a collection of agents cooperates to maximize the likelihood of visiting a certain state. In this example, four agents coordinate to control an ant-like robot; each agent controls one leg (2 joints/leg). The goal is to coordinate so that the ant moves to a specific position ({\color{purple}$\mathbf{\times}$}). No rewards are given; no distance metrics are required.
    }
    \label{fig:problem}
\end{wrapfigure}

Reinforcement learning (RL) has the potential to find novel strategies that solve complex tasks. From controlling fleets of autonomous vehicles to swarms of drones~\citep{cao_overview_2012, baldazo_decentralized_2019}, cooperative multi-agent RL~\citep{ippo, dec_pomdp} has the potential to find strategies that are more robust, scalable, and efficient than single-agent strategies.
However, finding novel strategies requires that human users do not pigeonhole agents into known solutions with dense rewards. Learning from sparse rewards remains a key open problem in multi-agent RL. \looseness=-1

In the area of single-agent RL, prior work has demonstrated that agents can learn from sparse rewards --- or no reward at all --- by leveraging self-supervised techniques~\citep{shelhamer2016loss, achiam2018variational, touati2022does, ghosh2019learning, eysenbach2018diversity,  emergent_exploration}. Goal-reaching is a canonical example: a human user provides the agent with a goal observation, and the agent attempts to reach that goal as quickly as possible~\citep {kaelbling1993learning, boyan1994generalization, dayan1992feudal, dietterich1998maxq, sutton1995generalization}. As the agent receives only a sparse reward upon reaching the goal, it is free to explore and experiment with different strategies for reaching the goal, including strategies that the human designer may not have envisioned. Importantly, prior work in this area has used self-supervised learning to make such sparse reward problems tractable~\citep{ding2019goal, kaelbling1993learning, lin2019reinforcement, sun2019policy}.

The main aim of this paper is to study how self-supervised techniques enable groups of agents to explore and cooperate to reach goals. We will focus on the problem of goal-conditioned multi-agent RL, where a single observation of a desired collective outcome specifies the task.
We propose a simple algorithm wherein each agent independently uses temporal contrastive learning to greedily select goal-oriented actions.
We find that this simple algorithm explores different coordination strategies over the course of learning, even in settings such as the StarCraft II Multi-Agent Challenge ($\mathtt{2s3z}$, $\mathtt{8m}$, $\mathtt{6h\_v\_8z}$, and $\mathtt{3s\_v\_5z}$), where prior methods never observe a single success. 
While our aim is not to claim that this method will always outperform well-engineered implementations of MARL algorithms that leverage reward functions, we do observe that this self-supervised approach is the only method to get nonzero reward in four environments, and almost triples the win-rate in the fifth.
Taken together, we believe this algorithm shows complex coordinated behaviors such as exploration, specialization, role differentiation can emerge from a simple learning rule and task specification (a single goal state), without any explicit mechanisms to produce these behaviors.

\section{Related Work}

Our work builds on prior work in goal-conditioned RL and independent learning for multi-agent RL~\citep{claus_dynamics_nodate}. In contrast to previous work in unsupervised and sparse-reward MARL, the bulk of which has focused on task-agnostic skill learning \citep{spd, hmasd} and explicit exploration mechanisms \citep{lagma, maser, cmae}, our method learns and explores in a fully end-to-end fashion.

\noindent\textbf{Independent Learning (IL) in MARL.}
Multi-agent reinforcement learning presents a fundamental choice for training: should agents learn together or separately?
Centralized methods like COMA and QMIX give agents access to the full environment state and let them share their policies and experiences~\citep{claus_dynamics_nodate, foerster_counterfactual_2024, rashid_qmix_2018}, executing in a decentralized manner during evaluation. Our approach builds off of independent learning~\citep{tan_multi-agent_1993} (e.g., IPPO~\citep{ippo}), where agents develop policies by training a shared-parameter policy on {\it individual} agent states and actions and collectively-achieved rewards. Each agent typically sees only parts of the environment. Prior work has shown that Independent PPO outperforms the fully centralized Multi-Agent PPO on complex StarCraft II benchmarks ~\citep{surprising_ppo} and scales better to larger environments~\citep{ippo}.

\noindent\textbf{Sparse Reward Methods in MARL.} 
Prior sparse reward methods in MARL typically use intrinsic motivation and domain-guided search to generate interesting agent behavior in the absence of an explicit reward signal \citep{maser, lazy, cmae, maven, formation-aware, subspace-aware}. MASER, for example, generates sub-goals for {\it individual} agents from a replay buffer~\citep{maser}. CMAE instead commands {\it collective} goals by searching for infrequently visited states in projected space to encourage exploration
\citep{cmae}. LAIES provides intrinsic motivation for causally-meaningful actions, defined using domain knowledge in the cooperative setting \citep{lazy}. More recently, methods have maximized diversity between successive joint policies \citep{JPD} or maintained a set of joint policies that effectively spans large regions of the environment \citep{population-based}. Our method will differ by not requiring domain-specific knowledge, subgoals, or explicit intrinsic motivation rewards.

One prior work, LAGMA \citep{lagma}, uses goal-conditioned trajectories as an intermediate step when maximizing an extrinsic reward. Whereas LAGMA relies on goal-guided search only after encountering reward signal, our method makes progress toward the goal in its absence, enabling us to operate in a strictly sparser binary reward regime. Furthermore, we learn to reach specified goal states directly, rather than using goals to shape an intrinsic reward alongside the extrinsic one. Finally, we target a single goal state instead of following an entire reference trajectory, which lets our method discover diverse paths to the goal during training.

\noindent\textbf{Goal-Conditioned Reinforcement Learning.}
Our work builds on a long line of prior goal-conditioned RL (GCRL) research~\citep{newell1959report, kaelbling1993learning, ghosh2019learning}, wherein a reinforcement learning agent attempts to reach a commanded goal state.
While sparse, the goal-conditioned setting is appealing from a user's perspective because it lifts much of the burden of reward function design~\citep{hadfield2017inverse, dulac2019challenges}: instead of hand-designing and implementing a reward function, a user simply gives one example of the desired outcome.
Prior self-supervised techniques for goal-reaching can tackle long-horizon sparse-reward problems~\citep{lin2019reinforcement, eysenbach2020c, chen2021decision, crl, andrychowicz2017hindsight, liu_single_2024}.
Our work extends these self-supervised techniques and observations to the multi-agent setting.

\section{Multi-Agent RL as Goal-Reaching} %
\label{sec:reframe}

This section introduces the formal definition of the multi-agent goal-reaching problem after reviewing the standard MARL problem.

\subsection{Preliminaries: Multi-Agent RL}

We consider a multi-agent RL problem with $n$ agents. At each timestep $t$, each agent $(i)$ receives a local observation $o_t^{(i)}$ and outputs an action $a_t^{(i)}$. Let  $s_t^{(i)}$ denote the state of agent $i$, with observations generated according to an agent-specific observation function $o_t^{(i)} \sim O_i(\cdot \mid s_t^{(i)})$. The environment transitions according to the stochastic transition function $p(s_{t+1}^{(1:N)} \mid s_t, a_t^{(1:N)})$ with initial state distribution $p_0(s_0^{(1:N)})$. At each timestep, the agents collectively receive a reward $r(o_t^{(1:N)}, a_t^{(1:N)})$. The overall objective is to maximize the expected discounted sum of these rewards:
\begin{equation}
     \E \left[ \sum_{t=0}^\infty \gamma^t r(o_t^{(1:N)}, a_t^{(1:N)}) \right]. \label{eq:marl}
\end{equation}

Following the IPPO paper~\citep{ippo}, we will treat each agent as policy operating on independent, local observations $\pi(a_t^{(i)} \mid o_t^{(i)})$. The parameters are shared across the agents (i.e., each agent has an identical policy). We define a local Q-function for each agent:
\begin{equation}
    Q(o_t^{(i)}, a_t^{(i)}) \triangleq \E\left[ \sum_{t' = t}^\infty \gamma^{t' - t} r(o_t^{(1:N)}, a_t^{(1:N)}) \Large \mid  o_t^{(i)}, a_t^{(i)} \right].
    \label{eq:local_Q}
\end{equation}
Here, the expectation is taken over the future actions of \emph{all} agents. Our analysis below will make use of the discounted state occupancy measure \citep{crl, puterman2014markov, liu_single_2024, ho2016generative}:
\begin{align}
    \rho^\pi_{\gamma}(s_f) \triangleq (1 - \gamma) \sum_{t=0}^\infty \gamma^t p_t^\pi(s_t = s_{f}),
    \label{eq:dsom}
\end{align}
where $p_t^\pi(s_t = s_{f})$ represents the probability of the agent being in state $s_{f}$ at time $t$.

\subsection{Defining the Multi-Agent Goal-Reaching Problem}
We now define the multi-agent goal-reaching problem, building on the Dec-POMDP formalism from prior work \citep{dec_pomdp, ippo, spd, lagma} and the GCRL framework \citep{kaelbling1993learning, newell1959report, ghosh2019learning}. Whereas the Dec-POMDP is typically defined in terms of a reward function, we will omit the rewards and instead include a space of goals.

We start by introducing the goals. Let $\gG$ be a space of goals and let $m_g^{(1:N)}: \gO^{(1:N)} \rightarrow \gG$ be a mapping from the observation space of all agents to the collective goal space. Goals are defined in this way because the objective is typically not to reach a particular state, but rather to reach {\it any} state that satisfies a desired property (e.g., any of the many possible states where the agents have successfully shot a basketball into a hoop). Let $p_g(g)$ denote the distribution over goals used for data collection and evaluation. Many of our experiments will use $p_g(g) = \delta(g = g^*)$, a Dirac distribution at one particular goal of interest (e.g., when all opponents have been defeated). Crucially, this choice of goal distribution eliminates the need to pre-define or adapt a goal curriculum, as done in prior work \citep{cmae, maser, lagma}.

As in the GCRL framework, the optimization objective is to maximize the probability of reaching the commanded goal, where the goal is a function of {\it all} observations $g_t = m_{g}^{(1:N)}(o_t^{(1:N)})$. Note that such mappings from full observations to relevant subsets of the observation for the goal are standard in problem settings from various prior works \citep{lin2019reinforcement, jaxgcrl, cmae}. We show how our method is not overly sensitive to relaxing the need for $m_g$ in \cref{sec:spec_of_mg}. We also discuss how many realistic MARL tasks meaningfully define goals in \cref{sec:meaningful_goals}.

We distinguish between a \textit{commanded} goal $g \in \gG$ and \textit{achieved} goal $g_t \in \gG$ as follows: a commanded goal is a desired goal space element that {\it conditions} a policy, while an achieved goal is where the agents currently are in the goal-space. The policy should eventually learn to reach $g_t = g$: the achieved goal should match the commanded goal.

To cast the goal-reaching setting as a reinforcement learning problem, we define the reward as
\begin{equation}
    r(o_t^{(1:N)}, a_t^{(1:N)}) = \begin{cases}
    1 & \text{if }g_t^{(1:N)} = g \\ 0 & \text{otherwise}
    \end{cases}.
\end{equation}
As it is currently stated, this reward definition does not make sense in settings with continuous states, as hitting the goal would be a measure-0 event. Thus, for generality, we define the reward function as the likelihood of hitting the goal at the \emph{next} time step:
\begin{equation}
    r(o_t^{(1:N)}, a_t^{(1:N)})
    = p(g_{t+1}^{(1:N)}= g \mid o_t^{(1:N)}, a_t^{(1:N)}). \label{eq:reward}
\end{equation}
These two reward functions are equivalent in expectation and thus result in equivalent optimization objectives (Eq.~\ref{eq:marl}) ~\citep{crl}. Finally, the maximized policy is a function of independent observations and a collectively desired goal, $\pi(a^{{(i)}}\mid o^{(i)}, g)$, following the inuition in the IPPO method \cite{ippo}.
In summary, the overall objective is:
\begin{align}
    \max_{\pi(a^{(i)} \mid o^{(i)}, g)}\underbrace{\mathbb{E}_{\substack{p_g(g)\\ \pi(\tau^{(1:N)} \mid g)}}\left[ \sum_{t=0}^{\infty} \gamma^t r(o_t^{(1:N)}, a_t^{(1:N)}) \right]}_{= \mathbb{E}_{p_g(g)} [\rho_{\gamma}^{\pi}(g)]}
    \label{eq:full_objective}
\end{align}
where $\tau^{(1:N)}$ is the sequence of observations and actions seen by all agents, $\pi(\tau^{(1:N)} | g)$ is the probability of sampling such a sequence given policy $\pi(a^{(i)} \mid o^{(i)}, g)$ and goal $g$, $\rho_{\gamma}^{\pi}(g)$ is the discounted state occupancy measure (Eq. \ref{eq:dsom}), and $p_g (g)$ is the distribution of goals. Intuitively, this objective corresponds to maximizing the time spent in the goal $g$. We note that using such a sparse, task-specific reward function is not new in MARL, and has been previously considered in a non-goal-conditioned setting \citep{cmae, subspace-aware, population-based}. However, the approach for solving this problem, which we present in the next section, differs from prior work by avoiding additional goal search, goal sampling, or use of task-specific goal knowledge \citep{maser, lazy, subspace-aware}.

\section{Self-Supervised Learning in the Multi-Agent Setting}
\label{sec:method}

We present an actor-critic method for multi-agent goal reaching building on contrastive RL ~\citep{crl}.
Following prior work~\citep{ippo, tan_multi-agent_1993}, we will address the multi-agent setting by learning decentralized policies and Q-functions. We deviate from strict independent learning in that we employ collective goals when learning the critic (although we can easily modify to use local goals as well without significantly hurting downstream performance).

\paragraph{Critic objective.}
We will use a variant of contrastive RL (CRL)~\citep{crl} to learn the Q-function, which is defined using a sparse goal-reaching reward (Eq. \ref{eq:reward}). 
CRL is a goal-conditioned RL (GCRL) method that uses a temporal contrastive objective to learn representations without requiring an external reward signal, traditionally defined in single-agent settings. The core of this method consists of two learnable encoders $\phi(o,a)$ and $\psi(g)$ that capture control-relevant temporal correlations between $(o,a)$ and $g$, trained by learning a classifier between actually reached future goals $g$ and random goals. At optimum, a learned energy function $f_{\phi, \psi}(o,a,g)$ over the representations parameterize the conditional discounted state occupancy measure $\rho^{\pi}(g \mid o, a)$ (Eq. \ref{eq:full_objective}); this is the single-agent critic in the goal-conditioned setting. Intuitively, these representations should enable agents to perform exploration directed toward a goal even when that goal has never been achieved, if the compressed representations learn generalized relationships between state-actions and goals. The resulting (scaled) Q-function can be modeled as the exponential of the distance between two representations:
\par \vspace{-2\parskip}\begin{equation*}
    f_{\phi, \psi}(o_t, a_t, g) = -\|\phi(o_t, a_t) - \psi(g)\|_2.
    \label{eq:q_val}
\end{equation*}

Extending CRL from the single-agent to the multi-agent setting can involve a variety of sampling methods. The multi-agent setting involves multiple interacting agents' states, observations, actions, and goals, and admits different possibilities for trained classifiers. 

In this case, we learn an energy function $f_{\phi, \psi}(o^{(i)},a^{(i)},g)$ that captures the temporal correlations between an {\it individual} agent's observations and actions $o^{(i)}, a^{(i)}$ with a collectively achieved goal in the future $g$. This energy function allows us to extract a value function appropriate for the local-observation, collective goal/reward and shared parameter setting following IPPO \cite{ippo}: individual agents can roll out actions from their individual observations, in service of reaching a collectively-desired goal.

We now walk through the precise objectives used to learn \ref{eq:q_val}. We learn these representations by optimizing a modified symmetric InfoNCE loss ~\citep{jaxgcrl, radford2021learning}. InfoNCE reformulates the problem of learning relative probabilities as a classification task between different distributions. Here, we specifically choose to sample positive {\it observation-action pairs of individual agents} and {\it achieved collective goals}. Negative samples are drawn uniformly at random over the collectively-achieved goals. In other words, a trained critic $f_{\phi, \psi}(o_t^{(i)}, a_t^{(i)}, g)$ prioritizes learning features that help individual agents distinguish achieved collective goal states actually encountered in the future from randomly chosen collective goal states. 

The symmetric InfoNCE loss for the multi-agent setting (Eq.~\ref{eq:critic}) relies on a batches  of $K$ samples \[\mathcal{B} = \{(o_1^{(i_1)}, a_1^{(i_1)}, \dots, (o_K^{(i_K)}, a_K^{(i_K)}, g_K) \}\] 
with pairs \textcolor{teal}{$((o_{i}^{(\alpha_i)}, a_{i}^{(\alpha_i
)}), g_{i} = m_g (o^{(1:N)}_{i+}))$} as positive examples and pairs \textcolor{teal}{$(o_{i}^{(\alpha_i)}, a_{i}^{(\alpha_i
})$}, \textcolor{purple}{$g_{j}$} and \textcolor{purple}{$(o_{j}^{(\alpha_i)}, a_{j}^{(\alpha_j
})$}, \textcolor{teal}{$g_{i}$} as negative examples, where $g_{
i
}$ are future achieved collective goals encountered after agent $\alpha_i$ encounters and acts on $(o_{i}^{(\alpha_i)}, a_{i}^{(\alpha_i)})$. In practice, we sample future time index $i_+ = i + \Delta i$ using $\Delta i \sim \text{Geom}(1-\gamma)$ following \ref{eq:dsom}.

Dropping the indexing for visual simplicity, upon convergence, the classifier $C_{\phi, \psi}((o,a), g) = \log \left(\frac{\exp(f_{\phi, \psi}(\textcolor{teal}{o,a,g}))}{\sum \exp(f_{\phi, \psi}(\textcolor{teal}{o,a},\textcolor{purple}{g}))}\right)$ learns temporal relationships between the underlying $(o,a)$ and $g$. The full symmetric InfoNCE loss is as follows:
\par \vspace{-2\parskip}{\footnotesize\begin{align}
    \min_{\phi, \psi}\mathbb{E}_{\mathcal{B}}\Bigg[&-\sum\nolimits_{\textcolor{teal}{i=1}}^{|\mathcal{B}|} \log \biggl( {\frac{e^{f_{\phi,\psi}(\textcolor{teal}{o_{i}}, \textcolor{teal}{a_{i}}, \textcolor{teal}{g_{i}})}}{\sum\nolimits_{\textcolor{purple}{j=1}}^{K} e^{f_{\phi,\psi}(\textcolor{teal}{o_{i}}, \textcolor{teal}{a_{i}}, \textcolor{purple}{g_{j}})}}} \biggr) - \sum\nolimits_{\textcolor{teal}{i=1}}^{|\mathcal{B}|} \log \biggl( {\frac{e^{f_{\phi,\psi}(\textcolor{teal}{o_{i}}, \textcolor{teal}{a_{i}}, \textcolor{teal}{g_{i}})}}{\sum\nolimits_{\textcolor{purple}{j=1}}^{K} e^{f_{\phi,\psi}(\textcolor{purple}{o_{j}}, \textcolor{purple}{a_{j}}, \textcolor{teal}{g_{i}})}}} \biggr) + 0.01 \cdot R(\phi, \psi)\Bigg], \label{eq:critic}
\end{align}}
where
\vspace{-1.0em}
\par \vspace{-2\parskip}{\footnotesize\begin{align}
    R(\phi, \psi) \triangleq & \log \left( {\sum\nolimits_{\textcolor{purple}{j=1}}^{K} e^{f_{\phi,\psi}(\textcolor{teal}{o_{i}}, \textcolor{teal}{a_{i}}, \textcolor{purple}{g_{j}})}} \right)  + \log \left( \sum\nolimits_{\textcolor{purple}{j=1}}^{K} e^{f_{\phi,\psi}(\textcolor{purple}{o_{j}}, \textcolor{purple}{a_{j}}, \textcolor{teal}{g_{i}})} \right).
\end{align}}

Mathematically, the exponential distance between representations $f_{\phi, \psi}(o_t^{(i)}, a_t^{(i)}, g)$ converges to an independent Q-function for individual agents: \[\exp (f_{\phi, \psi}(o_t^{(i)}, a_t^{(i)}, g)) \propto \rho_{\gamma}^\pi(g_{t+} = g \mid o_{t}^{(i)}, a_{t}^{(i)}),\]
where $\rho_{\gamma}^{\pi}$ denotes the $\gamma$-discounted state occupancy of a collective goal conditioned on agent $i$ observing $o^{(i)}$ and taking action $a^{(i)}$ at time $t$.

\paragraph{Actor objective.}
We use a neural network policy $\pi_\theta(a_t^{(i)} \mid o_t^{(i)}, g)$ that takes {\it individual} observations $o_t^{i}$ and collective goal as input. We assume the agents are homogeneous, so the same policy is used for modeling all agents (i.e., we employ parameter sharing following IPPO~\citep{ippo}). Non-homogeneous tasks can be made homogeneous by including the agent index or type as part of the observation space. 

Following single-agent CRL methods \citep{crl}, we train the policy by maximizing the (marginal-weighted) expected critic over states and goals sampled from the replay buffer:
\begin{align}
    \pi^*(a^{(i)} \mid o^{(i)}, g) &= \argmax_{\textcolor{teal}{\pi}} \; \E_{\substack{(o_t^{(i)}, g) \sim \gB\\a_t^{(i)} \sim \textcolor{teal}{\pi(a_t^{(i)} \mid o_t^{(i)}, g)}}}[f_{\phi, \psi}(o_t^{(i)}, a_t^{(i)}, g)] \\
    &\approx \argmax_{\textcolor{teal}{\pi}} \E_{\substack{(o_t^{(i)}, g) \sim \gB\\a_t^{(i)} \sim \textcolor{teal}{\pi(a_t^{(i)} \mid o_t^{(i)}, g)}}} \log \rho_\gamma^\pi (g_{t+} = g \mid o_t^{(i)}, a_t^{(i)}).
    \label{eq:actor}
\end{align}
Thus, the policy should pick {\it individual actions} that maximize the probability of occupying a {\it collective goal state}, and is enabled by a correctly-learned classifier $f_{\phi, \psi}(o_t^{(i)}, a_t^{(i)}, g)$.

\paragraph{Hindsight Experience Relabeling.} A key method that allows CRL to learn off of a sparse goal-achieving signal is Hindsight Experience Relabeling \citep{hindsight_relabel}. Namely, while we only use the single commanded goal $g$ during \textit{rollout} (for which the agent may never have seen a positive example), we \textit{train} the critic and update the actor using actualized goals $g_t$. Thus, the samples of $g_t$ and $g_{t+}$ in the above objectives are all actualized goals: in effect, we are telling the agent ``after taking action $a^{(i)}$, the collective eventually reached $g_{t+}$, and thus this experience is good training signal for $\pi(a^{(i)} \mid s^{(i)}, g_{t+})$.'' This approach allows us to use arbitrary future ``goals'' as a proxy to learn to reach target goals $g_t = g$. Such approaches have been shown to lead to emergent exploration within the single-agent setting \citep{liu_single_2024, emergent_exploration}.

\begin{algorithm}[t]
    \caption{\textbf{Independent CRL} is an actor-critic algorithm for multi-agent goal reaching.}
    \label{alg:icrl}
    {\footnotesize 
    \begin{algorithmic}

                                \State Initialize policy $\pi_\theta(a_t^{(i)} \mid o_t^{(i)}, g)$, decentralized critic $f_{\phi, \psi}(s_t^{(i)}, a_t^{(i)}, g) = \|\phi(o_t^{(i)}, a_t^{(i)}) - \psi(g)\|_2$, replay buffer $\gB$.
    \While{not converged}
        \State Sample goal $g \sim p_g(g)$ from the commanded goal distribution $p_{g}(g) = \delta_{g}(g)$.
        \State Collect episode using (independent) policies $\pi(a_t^{(i)} \mid o_t^{(i)}, g)$.
        \State Store episode $\{o_0^{(1:N)}, a_0^{(1:N)}, o_1^{(1:N)}, a_1^{(1:N)}, \cdots\}$ in buffer $\gB$.
        \State Sample observations $o_t^{(i)}$, actions $a_t^{(i)}$, and achieved future collective goals $g = m_g (o^{(1:N)})$ from $\gB$.
        \State Update critic $f_{\phi, \psi}$ with temporal contrastive learning (Eq.~\ref{eq:critic}).
        \State Update policy $\pi_\theta(a_t^{(i)} \mid o_t^{(i)}, g_{t})$ to maximize the critic (Eq.~\ref{eq:actor}).
    \EndWhile
    \State \textbf{return} policy $\pi_\theta(a_t^{(i)} \mid s_t^{(i)}, g)$.

    \end{algorithmic}}
\end{algorithm}

\subsection{Algorithm Summary}

We now summarize a complete algorithm, \textsc{Independent CRL} (ICRL), and provide pseudocode in Alg.~\ref{alg:icrl}.
ICRL works in the online setting, alternating between collecting data and updating the actor and the critic. In order to update the actor and critic, a batch of state-action pairs are sampled from the replay buffer.
We implement ICRL on top of JaxGCRL~\citep{jaxgcrl}. Code to reproduce our experiments is available in a code repository.

\section{Experiments}
\label{sec:experiments}

In this section, we aim to answer the following questions with empirics:

\begin{itemize}[topsep=0pt, itemsep=1pt]
    
    \item[] \hspace{-1em}\textbf{(Q1)} Is efficient exploration possible in long-horizon, sparse-reward tasks?
    \item[] \hspace{-1em}\textbf{(Q2)} How does exploration emerge in ICRL to solve these sparse learning tasks?
    \item[] \hspace{-1em}\textbf{(Q3)} Can independent CRL methods scale to continuous control tasks?
    \item[] \hspace{-1em}\textbf{(Q4)} Can re-framing a goal-reaching task as multi-agent simplify the problem?
\end{itemize}

We present all results with $\pm 1 \sigma$ error bars and average smoothing (to display 200 data points). The experiments (5 seeds per baseline per experiment) required 0.5-3 hours (ICRL, IPPO, and MAPPO) and 16 hours (MASER and LAGMA) per seed on a Tesla V100 GPU (32 GB). All experiments were run on an internal cluster.
See Appendix \ref{sec:main_results_summary} for a summary of all experiments and Appendix \ref{sec:experimental_details} for additional details. For environments with discrete actions, we use a variant of the Gumbel-Softmax trick (see Appendix \ref{sec:handle_discrete}).

\begin{figure}[t]
    \centering
    \includegraphics[width=0.8\linewidth]{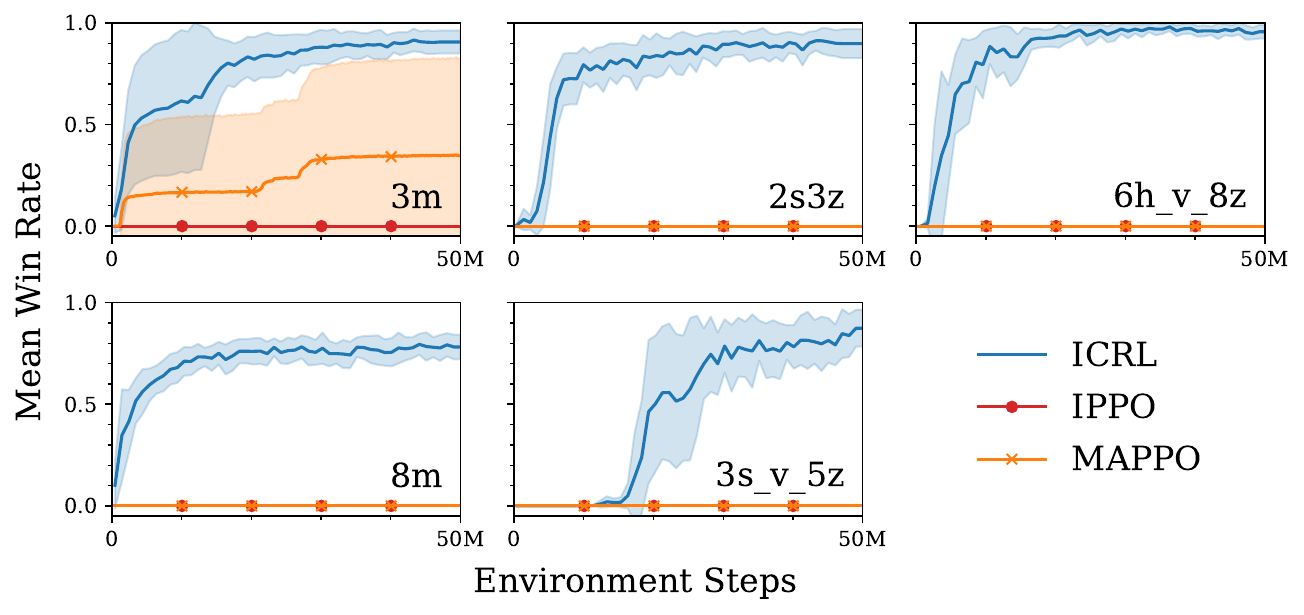}
    \caption{\footnotesize \textbf{Efficient learning on the StarCraft Multi-Agent Challenge (SMAX).} We compare ICRL (our method) to IPPO and MAPPO on five settings from the SMAX benchmark. On the \texttt{3m} setting, our method achieves a win rate that is $\sim 3\times$ higher than MAPPO, while the IPPO baseline has a win rate of zero. On the \texttt{2s3z}, \texttt{6h\_v\_8z}, \texttt{8m}, and \texttt{3s\_v\_5z} settings, only our method achieves a non-zero win rate.}
    \label{fig:SMAX_results} 
    \vspace{-1.5em}
\end{figure}

\subsection{Long-Horizon Cooperation}

After some initial sanity checks that the proposed method can solve simple variants of the game tag (see Appendix~\ref{sec:mpe_tag}) where the goal state is frequently observed, we turned our attention to long-horizon cooperation. It is intuitively unclear whether this method should work on long-horizon tasks where the goal state is only observed once after many (e.g., 50-100) steps, since reaching the goal involves extended blind exploration. Concretely, we evaluate the performance of ICRL on the StarCraft Multi-Agent Challenge (SMAC), a common benchmark in prior MARL work that pits a team of units against an enemy team \citep{smac}.
For our experiments, we use SMAX, a JAX implementation of a SMAC-like environment \citep{jaxmarl}. We test five classic SMAC environments: \texttt{3m}, \texttt{2s3z}, \texttt{6h\_v\_8z}, \texttt{8m}, \texttt{3s\_v\_5z}. We also test SMACv2 environments featuring randomized position and unit types (summarized in Fig.~\ref{fig:5R}).

As shown in Fig.~\ref{fig:SMAX_results}, \textbf{we find that ICRL consistently identifies successful policies early on in the training process}, despite the challenge of blind exploration. We later show qualitative evidence suggesting that agents are able to cope by developing coordinated intermediate strategies that behave non-trivially (Section ~\ref{sec:em_exploration}).

\begin{wrapfigure}[17]{r}{0.5\linewidth}
    \vspace{-0.7em}
    \centering
    \includegraphics[width=1.0\linewidth]{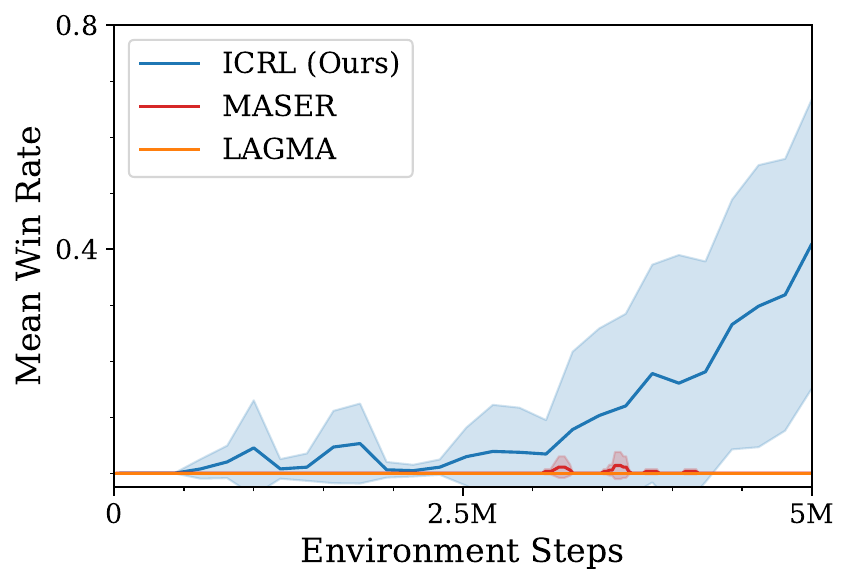}
    \vspace{-1.6em}
    \caption{\footnotesize ICRL (Ours) vs. MASER, LAGMA on SMAC (\texttt{2s3z}), using sparse rewards. Our method achieves 60\% win-rate by step 5M while both MASER's and LAGMA's win-rates are negligible.}
    \label{fig:MASER}
\end{wrapfigure}
To ensure reaching goals in sparse signal SMAC environments is challenging and requires nontrivial exploration, we compare ICRL with the multi-agent baselines IPPO and MAPPO, both of which have been previously found to be effective on a variety of SMAC environments \citep{surprising_ppo}. We also compare against more recent hierarchical methods MASER and LAGMA designed to learn with sparse rewards by decomposing tasks into subgoals ~\citep{maser, cmae, maven}. For a fair comparison, all methods receive sparse rewards (+1 if win, 0 otherwise); this is a sparser reward than presented in the original papers, which gives individual enemy reward bonuses and penalties for ally health loss. We find (Figure ~\ref{fig:SMAX_results} and ~\ref{fig:MASER}) that all these methods achieve near-zero success on most tasks, demonstrating the non-triviality of the goal-conditioned setting. A statistical significance test located in Appendix~\ref{sec:prob_improvment} shows that the probability of improvement for ICRL compared to MAPPO is on average 94\%. 

Concretely, we frame the goal as: reduce the sum of enemy healths to zero. Mathematically, this means that $m_g(o_t^{(1:N)})$ is the sum of the enemy healths and the goal $g$ is the scalar $0$. 
To ensure fair comparison with our goal-conditioned method, we provide the reward-driven baselines with a sparse reward (namely, a reward of 1 is given upon winning a battle).

\subsection{Emergent Exploration}
\label{sec:em_exploration}

As shown in previous experiments, our method performs well in long-horizon tasks where there is no signal until the end of the episode. For this to occur, agents must explore effectively in the interim before receiving any signal about which methods actually lead to the goal. How does the algorithm explore useful policies before observing even a single success? By visualizing learned policies at various points along training, we can see how exploration emerges in training and what, if any, unique skills are learned. 

We attribute this early-stage learning to emergent exploration, a phenomenon exhibited by many self-supervised RL algorithms. Previous work discussed emergent self-directed exploration in the single-agent setting, showing that giving a single goal leads to effective exploration \citep{liu_single_2024}. That work demonstrates contrastive representations are critical to learning useful skills before a single success is observed (a monolithic critic does not show directed exploration) \citep{liu_single_2024}. Recent concurrent work \cite{emergent_exploration} has made progress toward elucidating emergent exploration further and we summarize the key results here: learning a representation space which encodes the environment dynamics allows unsuccessful trajectories to be pruned from future exploration efforts. This allows agents to make progress toward the goal without explicit supervision through negative feedback. By extension to the multi-agent setting, we suspect that commanding a single goal to our method also enables such directed exploration. We empirically find that the learned curriculum of skills from ICRL surpasses subgoal-generating MASER, mirroring results in the single-agent setting~\citep{liu_single_2024}. We conclude, similar to previous work \citep{liu_single_2024}, that emergent exploration occurs under the following circumstances/conditions: \textit{(1)} environmental dynamics are learned in contrastive representations and \textit{(2)} a single long-horizon goal is commanded for CRL (rather than many human-picked subgoals).

\begin{figure*}[t]
    \centering
    \includegraphics[width=\linewidth]{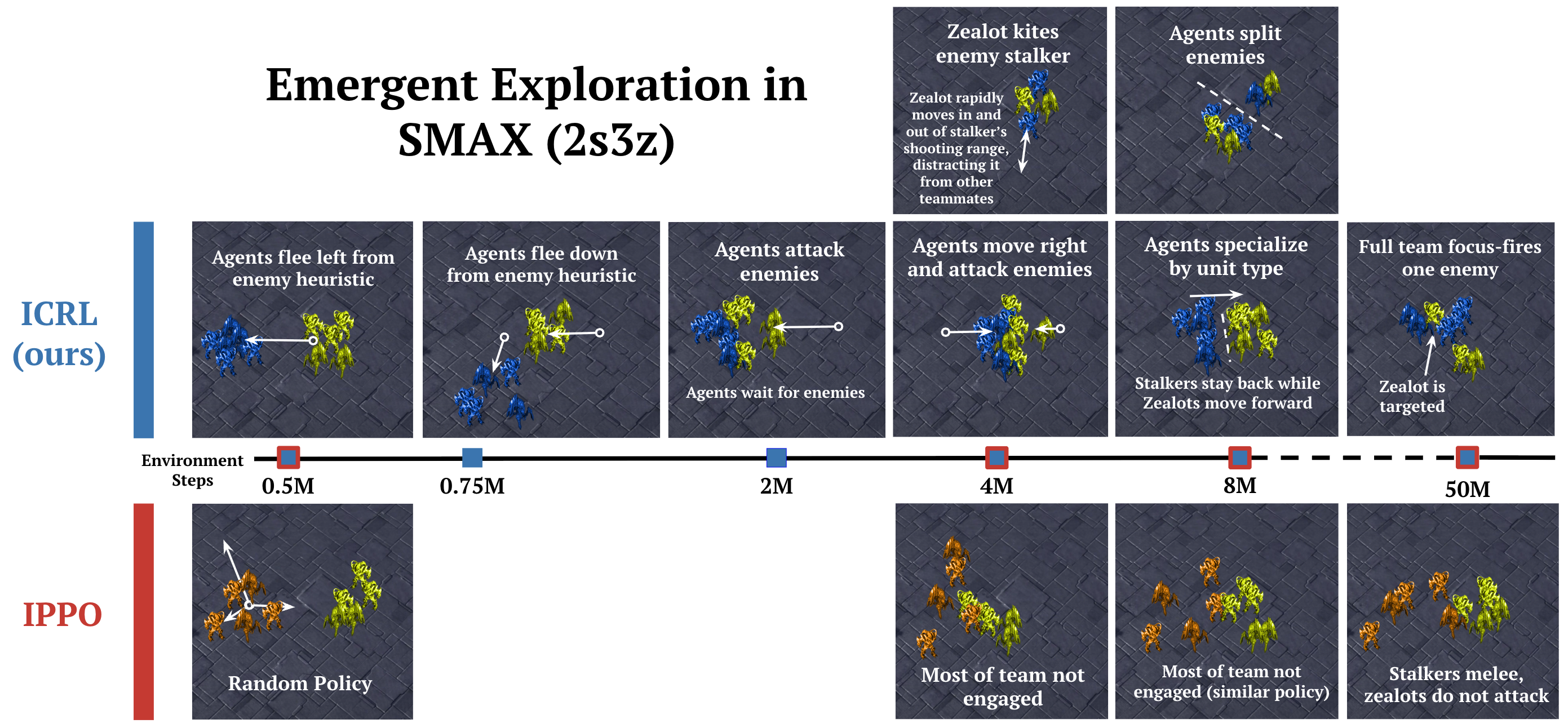}
    \caption{\footnotesize \textbf{How does ICRL learn to play StarCraft?} We visualize the exploration strategies of \emph{(Top Row)} ICRL (our method) and \emph{(Bottom Row)} IPPO on the SMAX (\texttt{2s3z}) environment over 50 million training environment steps. We observe that the ICRL algorithm explores different coordination strategies over the course of learning.}
    \label{fig:exploration_icrl}
    \vspace{-1.5em}
\end{figure*}

We use the \texttt{2s3z} environment (Figure \ref{fig:exploration_icrl}) to illustrate emergent exploration. \textbf{Our method learns basic skills (e.g., movement, attacking) early in training and slowly learns more advanced strategies, often well before observing a single success.} Initially, agents flee from the enemy units and hit the boundary, where they are eliminated. Later, agents learn to stay still and shoot the incoming enemy. By 2 million environment steps, before the agents have seen the goal-state even once, agents have explored various unique strategies. At 4 million steps, we observe common StarCraft unit micromanagement techniques: units learn unique skills such as ``kiting" and focus-fire \citep{jaxmarl}. At 8 million steps, agents learn more optimized flocking behaviors and learn to specialize by their unit type (we explore specialization more in Appendix \ref{sec:specialization}). Despite training with shared parameters, the policy network learns unique behaviors (for example, ranged attacks for the stalker or closer melee for the zealot unit). By the end of training, the agents not only retain the most successful skills, but also sync movements and focus-fire all at the same target.
This is in contrast to behaviors learned by non-goal conditioned methods such as IPPO, which generally cannot perform directed exploration without much signal. We observe that IPPO policies remain qualitatively similar and behave essentially randomly in benchmark training runs. These results support the view that a simple self-supervised rule with a minimal task specification (a single goal state) is sufficient to produce complex and coordinated exploratory behaviors.

\subsection{Tasks with Continuous Actions}
\label{mabrax_exp}

The tasks we have investigated so far have featured straightforward control mechanisms. Many real-world tasks, however, have an orthogonal dimension of complexity: manipulation of the agent in the physical world. We might expect ICRL to fail on such tasks: continuous control requires precise coordination across joints, which is challenging when each agent learns from local observations and a sparse goal signal alone.

Multi-Agent MuJuCo factorizes classic control tasks such as ant and half-cheetah, giving each agent control over only a subset of the available joints and observability of only the nearest other agents \citep{facmac}. Multi-Agent BRAX implements five of the Multi-Agent MuJuCo tasks \citep{jaxmarl}; we will use the Ant, Half-Cheetah, and Humanoid tasks (see Fig.~\ref{fig:problem}). We observe that our method performs well on this task and in Section \ref{sec:reframing} we explore how reframing even single-agent tasks as multi-agent increases performance.

\begin{wrapfigure}[15]{r}{0.5\linewidth}
    \centering
    \vspace{-1.2em}
    \includegraphics[width=1.0\linewidth]{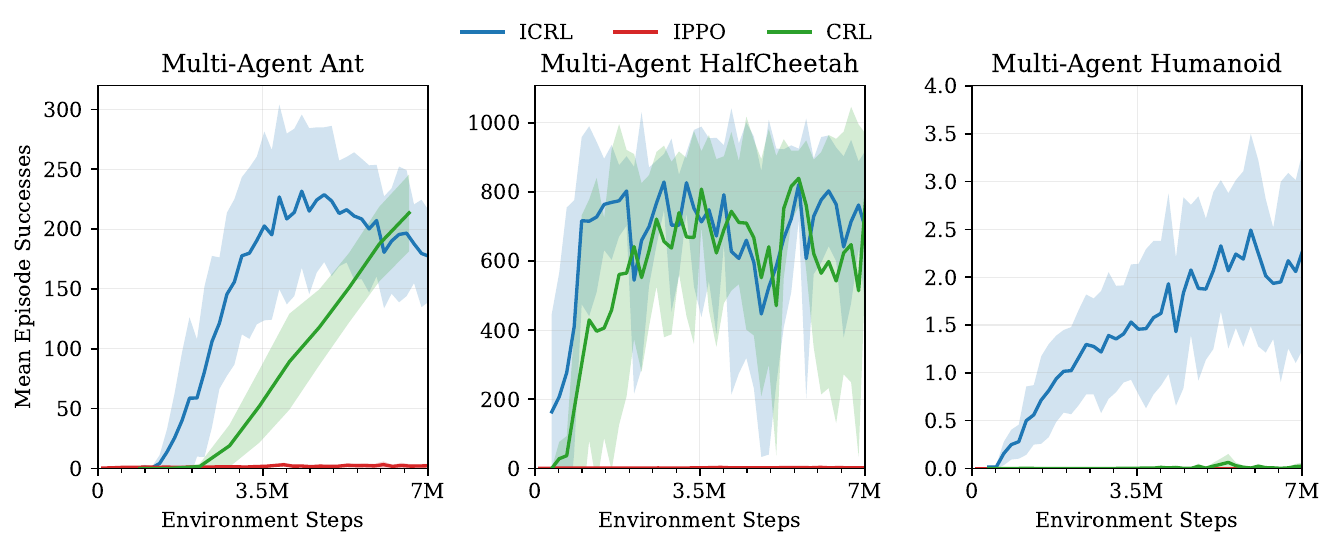}
    \caption{\footnotesize \textbf{Can ICRL solve continuous control tasks?} We compare ICRL (ours) to IPPO for controlling various robots where each limb's joints are controlled by a separate agent~\citep{jaxmarl}. IPPO makes no progress, perhaps because the sparse reward signal makes exploration challenging. Single-agent CRL is also compared, showing that casting even a single-agent problem as multi-agent allows the agent to learn faster.}
    \label{fig:MABRAX_results}
\end{wrapfigure}

We define the goal as controlling the robot so that its center of mass is at the desired goal. We define $m_g(o_t^{(1:N)})$ to extract the robot's $(x, y)$ position and sample goals $g \in \mathrm{R}^2$ uniformly on a disk of radius 10 meters. The reason we chose this specific distribution is that it is used by the JaxGCRL benchmark \citep{jaxgcrl} as the default goal distribution for the control task, making for a challenging but reasonable goal so that agents have to learn a non-trivial walking policy.
For evaluation purposes, a success is measured as a time-step (out of a total $1000$) when the agent is within $0.5$ meters of the goal.
While our method does not require rewards (it simply maximizes the likelihood of an event), the baselines use this success metric as a sparse reward.
The results, shown in Figure \ref{fig:MABRAX_results}, show that \textbf{our method is able to consistently reach the commanded goal on the sparse reward setting}. IPPO struggles to get non-zero reward, likely because the sparsity of the task makes exploration challenging.

\subsection{Reframing Single-Agent Problems as Multi-Agent Problems}
\label{sec:reframing}

Multi-agent reinforcement learning (MARL) is typically viewed as an extension of single-agent RL, with multi-agent tasks considered strictly harder versions of their single-agent counterparts. Here, we explore the opposite view: that MARL might be understood as making independence assumptions that allow us to learn more efficiently. These independence structures, which enable more data-efficient learning and better generalization in other areas of machine learning \citep{graphical_models}. We compare ICRL against single-agent CRL~\citep{crl} on various control tasks where CRL controls all joints with one policy while ICRL factors control across limbs (Figure~\ref{fig:MABRAX_results}).
We observe that learning separate policies to control limbs in a decentralized manner results in more sample-efficient learning and earlier success rates on two tasks, with one task also getting higher asymptotic success. One explanation for this potentially surprising finding is that factoring the robot into various independently-controlled agents actually reduces the hypothesis space: instead of searching over all policies involving multiple joints in all limbs, each agent can focus on controlling just the joints in its corresponding limb.

\section{Conclusion}

In this paper, we have studied how a simple self-supervised learning rule can induce coordination and exploration in goal-conditioned multi-agent settings. While this problem setting entails very sparse feedback, our experiments demonstrate that the proposed method can make progress on solving these tasks. We do not claim that our method is anywhere near the maximum possible performance with self-supervised techniques in general, but rather offer these experiments as evidence that multi-agent goal-reaching is a tractable problem statement. We believe this work takes a small step towards the ambitious goal of understanding how minimal algorithms and simple task definitions can explain the complex behavior observed in natural decision-making settings.

\paragraph{Limitations.}
Although formulating a problem as goal-reaching allows the task to be specified more easily via a single goal rather than a reward function, it may not always be clear how to specify certain tasks as goals. Similarly,  it is sometimes possible to encode one task in two different ways (choices of $\gG$ and $m_g$), yet result in slightly different learning behaviors (see Appendix \ref{sec:spec_of_mg}).

\newpage
\appendix

\section{Main Experimental Results}
\label{sec:main_results_summary}

\begin{restate}[Experimental Setup]
All experiments use sparse 0/1 rewards: +1 when in the goal state, 0 otherwise. For SMAX environments, the goal is to reduce enemy health to zero. For continuous control, the goals are spatial positions. We compare against IPPO, MAPPO, and MASER baselines using identical sparse reward signals.
\end{restate}

Table \ref{tab:experimental_summary} summarizes the main experimental results:

\begin{table}[h]
\centering
\begin{tabular}{lllc}
\toprule
\textbf{Environment (Metric)} & \textbf{Task} & \textbf{Approach} & \textbf{Result} \\
\midrule
\multirow{4}{*}{\shortstack[l]{\textbf{MPE Tag} \\ (Mean Episode Return)}}
    & \multirow{2}{*}{3 Agents}
        & \textbf{ICRL (Ours)} & \textbf{4704.62 ± 850.18} \\
    &   & IPPO \citep{ippo} & 3643.21 ± 82.01 \\
    \\
    & \multirow{2}{*}{6 Agents}
        & \textbf{ICRL} & \textbf{16293.59 ± 2204.55} \\
    &   & IPPO & 5158.75 ± 1759.75 \\
    
\midrule
\multirow{2}{*}{\shortstack[l]{\textbf{Multi-Agent Control} \\ (Mean Success Rate)}}
    & Ant & \textbf{ICRL (Ours)} & \textbf{270.63 ± 30.02} \\
    &     & IPPO & 6.94 ± 8.94 \\
    &     & CRL \citep{crl} & 235.33 ± 35.33 \\
    
    \\
     & Half-Cheetah & \textbf{ICRL} & \textbf{901.58 ± 12.10} \\
    &     & IPPO & 0.0 ± 0.0 \\
    &     & CRL & 902.62 ± 21.71 \\

    \\
     & Humanoid & \textbf{ICRL} & \textbf{3.38 ± 0.90
} \\
    &     & IPPO & 0.00 ± 0.00 \\
    &     & CRL & 0.22 ± 0.30 \\
    
\midrule

\multirow{16}{*}{\shortstack[l]{\textbf{StarCraft Multi-Agent} \\ \textbf{Challenge (SMAX)} \\ (Mean Win Rate)}}
    & \multirow{3}{*}{3m}
        & \textbf{ICRL (Ours)} & \textbf{0.94 ± 0.06} \\
    &   & IPPO & 0.00 ± 0.00 \\
    &   & MAPPO \citep{surprising_ppo} & 0.36 ± 0.49 \\
    
    \\
    & \multirow{4}{*}{2s3z}
        & \textbf{ICRL} & \textbf{0.95 ± 0.02} \\
    &   & IPPO & 0.00 ± 0.00 \\
    &   & MAPPO & 0.00 ± 0.00 \\
    &   & MASER \citep{maser} & 0.01 ± 0.02 \\
    &   & LAGMA \citep{lagma} & 0.00 ± 0.00 \\
    
    \\
    & \multirow{3}{*}{6h\_v\_8z}
        & \textbf{ICRL} & \textbf{1.00 ± 0.00} \\
    &   & IPPO & 0.00 ± 0.00 \\
    &   & MAPPO & 0.00 ± 0.00 \\
    
    \\
    & \multirow{3}{*}{8m}
        & \textbf{ICRL} & \textbf{0.84 ± 0.07} \\
    &   & IPPO & 0.00 ± 0.00 \\
    &   & MAPPO & 0.00 ± 0.00 \\
    
    \\
    & \multirow{3}{*}{3s\_v\_5z}
        & \textbf{ICRL} & \textbf{0.95 ± 0.03} \\
    &   & IPPO & 0.00 ± 0.00 \\
    &   & MAPPO & 0.00 ± 0.00 \\
\bottomrule
\\
\end{tabular}
\caption{Maximum performance across environments. We report episode returns for MPE Tag, success rate for Multi-Agent Control, and win-rate for SMAX tasks. All values are reported as the maximum result $\pm 1 \sigma$ at the timestep the maximum result is achieved.}
\label{tab:experimental_summary}
\end{table}

\subsection{Addressing Low Baseline Performance on Sparse Rewards}
To ensure fair comparisons in our goal-conditioned setting (where an agent only gets reward signal when in the goal state), we give all methods access to the same sparse rewards: $+1$ if in the goal state (e.g., win state in SMAX or the goal position in MABRAX Ant) and $0$ otherwise. 

The goal-conditioned 0/1 reward specification is a natural choice for any task-specific objective, where the successful completion of the task is a function of agent observations. Notably, the goal-conditioned reward does not need any domain-specific knowledge beyond the specification of the task itself. Unlike LAIES \citep{lazy}, we do not need to specify external states in fully cooperative settings, which would require domain-specific knowledge for different tasks. Furthermore, this 0/1 reward specification is not new and has been previously used to benchmark methods: other papers such as CMAE \citep{cmae} and LAIES \citep{lazy} also use such a setting for their method comparisons, and find similarly low performance in their tested baselines (though the compared methods are different).

To address the concern that the IPPO and MAPPO baselines are not representative of the current SOTA in sparse reward settings (such as the goal-conditioned setting), we directly compare our method to MASER \citep{maser}, a MARL algorithm designed for sparse reward that has shown superior performance to state-of-the-art MARL algorithms including QMIX \citep{rashid_qmix_2018}, MAVEN \citep{maven}, and COMA \citep{foerster_counterfactual_2024} without further signal from expert domain knowledge. ICRL outperforms MASER in the 0/1 reward setting (\cref{fig:MASER}). We believe at least one of the reasons for ICRL's high relative performance on such sparse-signal environments is the method's ability to explore effectively. This emergent exploration is discussed further in \cref{sec:em_exploration}. 

\section{Experimental Details}
\label{sec:experimental_details}

\begin{restate}[Independent CRL Method]
Our method treats each agent as an independent contrastive learner with shared parameters. The critic learns representations $\phi(o,a)$ and $\psi(g)$ using the symmetric InfoNCE loss, while the actor maximizes $\mathbb{E}[-\|\phi(o_t^{(i)}, a_t^{(i)}) - \psi(g)\|_2]$ to output actions that are close to the goal in representation space.
\end{restate}

Code and hyperparameters for reproducing all experiments can be found in a code repository.\footnote{\url{https://github.com/Chirayu-N/gc-marl}} We highlight some key hyperparameters in Table \ref{tab:mpe_hyperparams}.

\begin{table}[h]
    \centering
    \begin{tabular}{lc}
        \toprule
        \textbf{Hyperparameter} & \textbf{Value} \\
        \midrule
        Total Environment Steps & 50,000,000 \\
        \# Epochs & 500 \\
        \# Environments & 256 (64) \\
        \# Eval Environments & 64 \\
        \midrule
        Actor LR & 3e-4 \\
        Critic LR & 3e-4 \\
        Alpha LR & 3e-4 \\
        \midrule
        Batch Size & 256 (64) \\
        Gamma & 0.99 \\
        LogSumExp Penalty Coefficient & 0.1 \\
        \midrule
        Max Replay Size & 5,000 \\
        Min Replay Size & 1,000 \\
        Unroll Length & 62 \\
        \bottomrule
        \hspace{1em}
    \end{tabular}
    \caption{Independent CRL Hyperparameter Values for the MPE Tag, MABRAX, and SMAX environments. Note for the 6h\_v\_8z SMAX environments, we needed to reduce environments and batch size (listed in parentheses) to avoid out-of-memory errors.}
    \label{tab:mpe_hyperparams}
\end{table}

We found our method to be fairly robust: we did not perform any hyperparameter tuning across the various environments. One exception was for the larger SMAX environments, where we reduced the number of environments and batch size to avoid out-of-memory errors. For all experiments, we test all algorithms without RNNs. For our SMAX experiments, we use the available action mask for all algorithms.

\subsection{Baseline Hyperparameters}
To the best of our knowledge, the hyperparameters from the JAXMARL paper are tuned to each task (see Appendix G of the JAXMARL paper \cite{jaxmarl}, where each hyperparameter is taken either from the original paper or new hyperparameters specifically for the JAX task if performance is better). Furthermore, performance of JAXMARL is “thoroughly benchmark[ed]” on correctness to ensure “equivalent agent performance” \cite{jaxmarl}. 

Note again that we did not hyperparameter-tune our method at all, and highlight that our results are for the sparse-reward setting.

\section{Handling Discrete Actions}
\label{sec:handle_discrete}
In our experiments, we use both environments with continuous actions and environments with discrete actions. For the discrete action tasks, we train the policy network with the Straight-Through Gumbel-Softmax trick~\citep{jang2016categorical, maddison2016concrete} for backpropagating gradients through the sampling of the discrete action. We deviate from prior work by parameterizing our critic as a function of the soft actor output rather than the discrete action itself. Please refer to the code block below for further clarification.

We ran the SMAX \texttt{2s3z} experiment with and without our modification to the critic parametrization, and found that giving the one-hot actions to the critic results in poor performance, with zero successes throughout the training run. We suspect this discrepancy is because critic representations are crucial for early-stage exploration and learning, and the critic’s soft action parametrization allows us to compute exact gradients through the critic during actor loss instead of approximate gradients as is usually done.

\begin{algorithm}
\begin{lstlisting}
hard_actions = jax.nn.one_hot(jax.nn.argmax(logits), num_actions)
soft_actions = jax.nn.softmax(logits, axis=-1)

# CRITIC LOSS
loss = critic_loss(obs, soft_actions, achieved_goals)    # ours
# loss = critic_loss(obs, hard_actions, achieved_goals)  # conventional

# ACTOR LOSS
loss = f(obs, soft_actions, achieved_goals)              # ours
# loss = f(obs, (hard_actions - soft_actions).detach() +
#        soft_actions, achieved_goals)                   # conventional

\end{lstlisting}
\end{algorithm}

\section{Defining Meaningful Goals}
\label{sec:meaningful_goals}
The focus of our paper is on goal-reaching tasks. Nonetheless, we believe that many tasks can be roughly cast as goal-reaching. For example, a foraging task could be formulated as trying to collect (say) 10 trees; it also might be specified as collecting more trees than another player (i.e., the goal is a binary indicator indicating which player has collected more trees).

We note that the rest of the JAXMARL \cite{jaxmarl} benchmark tasks, none of which were explicitly designed for goal-reaching, can all be roughly cast as goal-reaching:

\begin{table}[h]
    \centering
    \begin{tabular}{lc}
        \toprule
        \textbf{Task} & \textbf{Goal} \\
        \midrule
        Simple World Comm (good agent) & set vector [food eaten, distance to observed predators] to [$K$, $K$] \\
        Simple World Comm (adversary) & same as MPE Tag (see paper section \ref{sec:mpe_tag}) \\
        Overcooked & set number of pending recipes to zero OR number of soups done to $K$ \\
        Hanabi & form five consecutively ordered stacks (explicit goal of the game) \\
        JaxNav & set self’s position to the destination \\
        STORM & set payoff to $K$ \\
        Coin Game & set payoff to $K$ \\
        Switch Riddle & set payoff to $K$ \\
        \bottomrule
        \hspace{1em}
    \end{tabular}
    \caption{Examples of meaningful goals defined for a every JAXMARL \cite{jaxmarl} tasks. $K$ represents an arbitrarily large value or maximum possible payoff if known analytically.}
    \label{tab:mpe_hyperparams}
\end{table}

\vspace{-1.5em}
\section{Additional Experiments}

\subsection{Didactic Experiment: Teamwork on MPE Tag}
\label{sec:mpe_tag}
Our first experiment aims to study whether our algorithm ICRL works at all. To test this, we choose a simple task where prior methods are known to work.

\begin{wrapfigure}[18]{r}{0.5\linewidth}
    \centering
    \vspace{-1em}
    \includegraphics[width=1.0\linewidth]{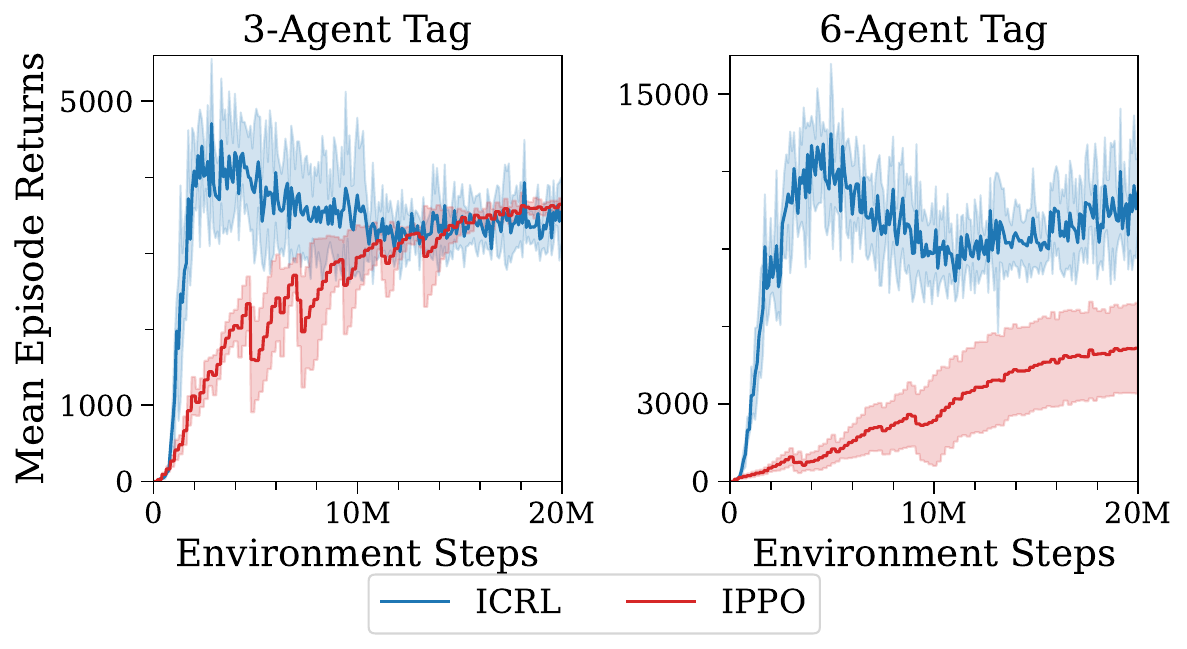}
    \caption{\footnotesize \textbf{``Tag'' as a goal-reaching problem.} We compare ICRL (ours) to IPPO on the Multi-Particle Agent Tag FACMAC Environment, including both the 3-agent setting \emph{(Left)} and the 6-agent setting \emph{(Right)}. We observe that our method learns faster in both settings and reaches higher asymptotic returns in the 6-agent setting.}
    \label{fig:mpe_tag_results}
\end{wrapfigure}

In the MPE Tag environment, predator agents must collide with (or “tag”) a faster prey. The environment is partially observed: if agents or landmarks are outside a fixed view radius, their observation gets masked with a placeholder value.  We use the FACMAC (Factored Multi-Agent Centralized Policy Gradients) variant, where the prey uses a simple heuristic policy, so the environment is a fully cooperative Dec-POMDP \citep{jaxmarl, facmac}. 
We frame the goal as: set the distance of the nearest (observed) predator to the prey to zero. Mathematically, we set $m_g(o_t^{(i)})$ as the distance between the closest predator and prey, and estimate the goal $g$ equal to the scalar $0$. If the prey is not visible, this is set to an arbitrarily large value. The results are summarized in Figure \ref{fig:mpe_tag_results}.

\begin{wrapfigure}[14]{r}{0.5\linewidth}
    \centering
    \vspace{-1.5em}
    \includegraphics[width=1.0\linewidth]{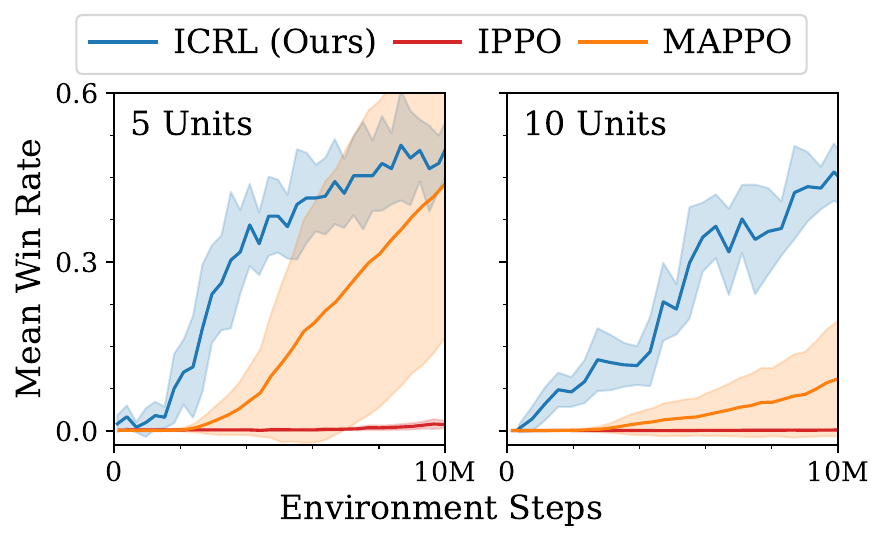}
    \caption{\footnotesize We compare ICRL (ours) to IPPO and MAPPO on the randomized 5-agent and 10-agent SMACv2 environments. Our approach learns faster than both baselines, achieving higher asymptotic returns than IPPO. IPPO's win rate is effectively zero.}
    \label{fig:5R}
\end{wrapfigure}

Our results demonstrate that ICRL consistently performs well against baseline methods. On the MPE Tag task, \textbf{ICRL matched or beat the performance of IPPO both qualitatively (via interesting formations and strategies) and quantitatively (gaining more rewards at its peak)}. While ICRL matched performance on the 3-agent environment, it exceeded the IPPO baseline on 6 agents, converging to an effective policy much more quickly.

\subsection{Performance on SMACv2 environments}

As shown in Figure \ref{fig:5R}, on the 5-agent and 10-agent SMACv2 environments, we find that our method can consistently learn how to coordinate to defeat the enemy, learning more quickly than both IPPO and MAPPO. While, for the 5-agent environment, MAPPO converges to a more successful policy on average, ICRL achieves a higher overall win-rate for the more challenging 10-agent version.

Interestingly, we find that the reward-driven algorithms better navigate the challenge of sparse reward learning on this environment than they do on the standard SMAC environments, even though the SMACv2 environments are generally thought to be more challenging.  While IPPO fails to get even a single success on \texttt{3m}, \texttt{2s3z}, or \texttt{6h\_v\_8z}, it gets a non-trivial success rate on the SMACv2 environment. 

We suspect that the inherent stochasticity present in this environment may naturally encourage agents to explore more widely, leading to the reward-based baselines performing better on these tasks than in the standard SMAC task. A statistical significance test located in Appendix~\ref{sec:prob_improvment} shows that the probability of improvement for ICRL compared to MAPPO is on average 94\%.

\subsection{Robustness to Not Specifying Goal-Space $\mathcal{G}$ and Goal-Mapping $m_g$}
\label{sec:spec_of_mg}

\begin{restate}[Multi-Agent Goal-Conditioned RL Problem]
We consider a multi-agent RL problem where agents cooperate to reach a commanded goal state $g \in \mathcal{G}$, with mapping $m_g: \mathcal{O}^{(1:N)} \rightarrow \mathcal{G}$ from observations to goals. The goal-conditioned reward is defined as (for continuous states): \[r(o_t^{(1:N)}, a_t^{(1:N)}) = P(g_{t+1} = g \mid o_t^{(1:N)}, a_t^{(1:N)}),\]
with the overall objective being:

\[
    \max_{\pi(a^{(i)} \mid o^{(i)}, g)}\mathbb{E}_{p_g(g), \pi(\tau^{(1:N)}|g)}\left[ \sum_{t=0}^{\infty} \gamma^t r(o_t^{(1:N}, a_t^{(1:N)}) \right]
\]
\end{restate}

In order to address the concern that picking a goal space $\mathcal{G}$ and mapping function $m_g$ requires user specification and, therefore, provides the method additional information, we test our method by using uninformative and non-task dependent choices of $\mathcal{G}$ and $m_g$. We run our ablation experiment for the SMAX \texttt{2s3z} environment. Relative to the other tested benchmarks, MPE Tag FACMAC and MABRAX, the SMAX implementation involves the most nontrivial $m_{g}$: the mapping computes the sum over enemy healths, as opposed to a simple observation truncation (MABRAX) or translation of task reward into goal (MPE Tag). Thus, we believe the results presented here generalize over to other benchmarks. 

\begin{figure}[htbp]
    \centering
    \vspace{-1em}
    \includegraphics[width=0.5\linewidth]{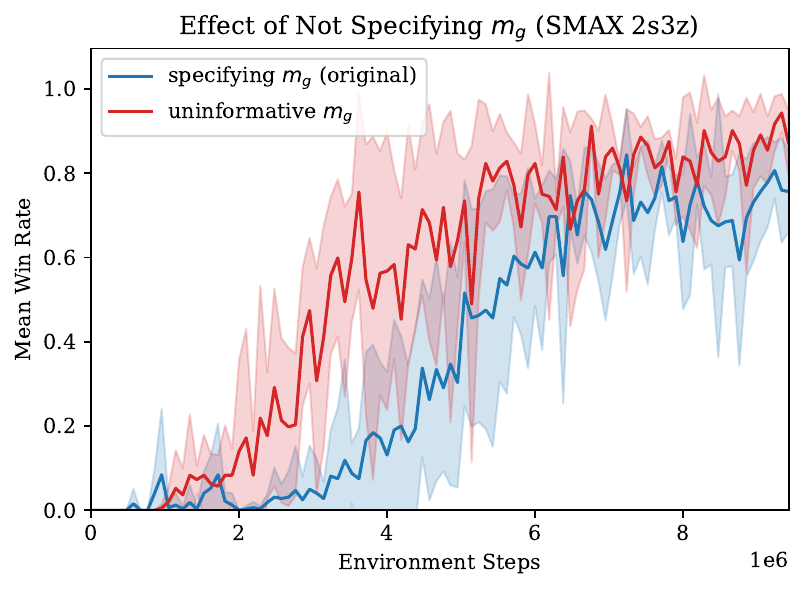}
    \caption{Specifying $m_g$ is not necessary for good performance. $\pm 1 \sigma$ error bars.
    }
    \label{fig:No_MG}
    \vspace{-1em}
\end{figure}

We let $\mathcal{G} = \mathcal{O}^{(1:N)}$ (the full observation space) and let $m_g = I$ (the identity map), and provide the method a single goal, commanding a collective state where every enemy health is 0 while filling out the remaining values using an arbitrary state from a previously collected trajectory. As seen in \cref{fig:No_MG}, ICRL achieves non-trivial performance on this task—in fact, our method performs even better than it did with a human-selected choice of $m_g$ that isolates the enemy's health. This shows that ICRL is robust to the choice of $m_g$ in a complex cooperative MARL benchmark and, in fact, may not need this specification at all to complete challenging tasks.

One hypothesis for why we observe these results is that the full observation provides additional contextual information that may be relevant to goal-reaching. While a manually-designed $m_g$ can help reduce dimensionality and focus learning on intuitive task-relevant observation features, it may inadvertently remove information that is useful in learning more complex environmental dynamics and coordination strategies.

Note that specifying $m_g$ naturally reflects that the goal is a {\it property} of observations, rather than a specific observation. For instance, although ICRL only receives a single example of the goal in this SMAX experiment, that example still overspecifies the goal: the real objective is to eliminate enemies, not to eliminate enemies {\it and} reach a randomly-specified state with goal-irrelevant observations. Thus, although the results in \cref{fig:No_MG} suggest $m_{g}$ is not necessary for ICRL's performance, we keep the goal-mapping function so that our formulation aligns with the nature of real tasks.

\subsection{Probability of Improvement}
\label{sec:prob_improvment}
In order to determine the statistical significance of our results, we use the probability of improvement, which is the chance that, in a randomly selected environment, a given algorithm performs better than another \citep{agarwal2022deepreinforcementlearningedge}. 

\begin{wrapfigure}[14]{r}{0.5\linewidth}
    \vspace{-1em}
    \centering
    \includegraphics[width=1.0\linewidth]{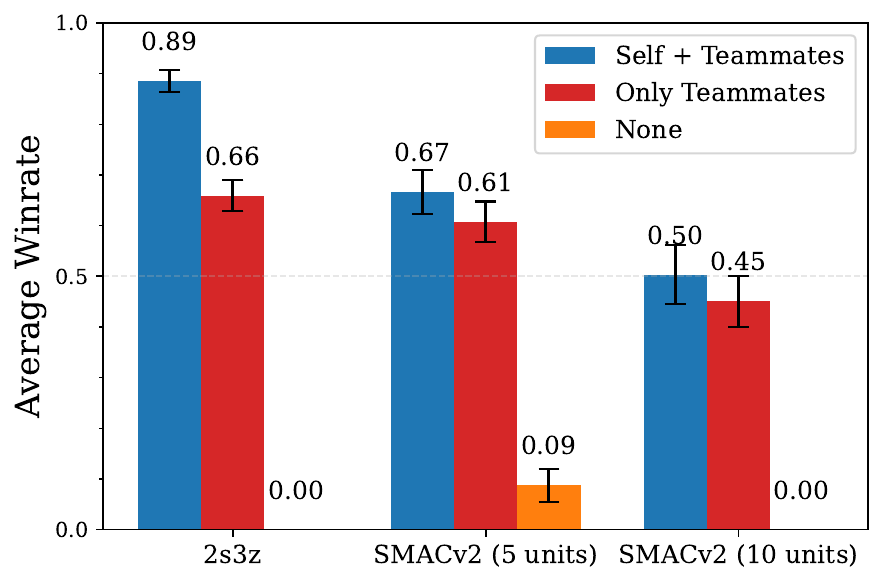}
    \caption{\footnotesize \textbf{Agents specialize in SMAC, using information about their player type to make decisions.}
    Win rates decline as unit type information is progressively removed from agent observations: baseline (self + teammate types), partial ablation (teammate types only), and full ablation (no type information).}
    \vspace{-2em}
    \label{fig:specialization}
\end{wrapfigure}
We compute the probability of improvement for ICRL compared to MAPPO over all SMAX environments using the max win rate across a training run as the measure of performance, and obtain an estimate of 0.94 with a 95\% bootstrapped confidence interval of $[0.85, 0.99]$. When using the final episode win rate instead, we obtain an estimate of 0.86 with a 95\% bootstrapped confidence interval of $[0.73, 0.95]$. In both cases, the probability of improvement compared to IPPO is 1.0.

\subsection{Do Agents Specialize?}
\label{sec:specialization}
Given that all agents use identical policy network architectures with unit type as the only distinguishing observation feature, one might expect uniform policy learning across different unit types. Yet prior research demonstrates that role specialization improves performance in cooperative multi-agent tasks \citep{smac, ippo, surprising_ppo}. Does our method learn to specialize?

To investigate this, we conduct an ablation study where we systematically remove unit type information from agent observations. We compare three conditions: agents observing both their own and teammates' unit types (baseline), agents observing only teammates' unit types, and agents observing no unit type information. If ICRL relies on unit type specialization, we expect performance to degrade as this information is removed. Figure \ref{fig:specialization} demonstrates this expected performance decline on \texttt{2s3z} and the SMACv2 5-unit and 10-unit environments (these are the only environments that include heterogeneous unit types). Interestingly, there is a significantly smaller drop in performance when removing an agent's own unit type from its observation, implying that information about the distribution of unit types across teammates is more relevant to task performance. The 0\% win rate on \texttt{2s3z} when removing all unit type information may seem counterintuitive, especially since the teammate unit type distribution does not provide extra information beyond the agent's unit type in this environment. We hypothesize that this outcome arises because the policy network was only exposed to a single team unit type configuration during training, and thus does not behave as expected under altered inputs.

\subsection{Baseline Comparison with Exploration Mechanisms}
\label{sec: MAPPO_exploration}

Since we partially attribute the effectiveness of our method to its ability to jointly explore without much guidance, we wanted to understand whether this effect could be reproduced using conventional exploration mechanisms in combination with standard MARL algorithms. In order to teset this we evaluated MAPPO (the most performant baseline on SMAX) on the 2s3z environment with the following mechanisms: action space noise, action entropy term in actor loss, parameter space noise \citep{plappert2018parameterspacenoiseexploration}.

We tried these modifications with 5 random seeds and swept over the relevant hyperparameters (standard deviation of Gaussian noise, entropy coefficient). We found that zero successes were achieved even with these modifications, suggesting that the contrastive learning scheme is crucial for this kind of exploration.

\subsection{Comparison with other Multi-Agent Hindsight Relabelling Approaches}

\begin{table}[h]
\centering
\begin{tabular}{cccccc}
\toprule
env step & ICRL & ICL(TD) & ICL(MC) & IGCBC & IHER \\
\midrule
0M & 0.0 ($\pm$0.0)   & 0.0 ($\pm$0.0)   & 0.0 ($\pm$0.0)    & 0.0 ($\pm$0.0)   & 0.0 ($\pm$0.0) \\
1M & 0.08 ($\pm$0.07) & 0.0 ($\pm$0.0)   & 0.0 ($\pm$0.0)    & 0.0 ($\pm$0.0)   & 0.0 ($\pm$0.0) \\
2M & 0.08 ($\pm$0.05) & 0.02 ($\pm$0.02) & 0.0 ($\pm$0.0)    & 0.0 ($\pm$0.0)   & 0.0 ($\pm$0.0) \\
3M & 0.03 ($\pm$0.03) & 0.02 ($\pm$0.02) & 0.03 ($\pm$0.03)  & 0.09 ($\pm$0.04) & 0.0 ($\pm$0.0) \\
4M & 0.18 ($\pm$0.09) & 0.01 ($\pm$0.01) & 0.05 ($\pm$0.04)  & 0.33 ($\pm$0.04) & 0.0 ($\pm$0.0) \\
5M & 0.33 ($\pm$0.11) & 0.05 ($\pm$0.04) & 0.125 ($\pm$0.07) & 0.31 ($\pm$0.05) & 0.0 ($\pm$0.0) \\
8M & 0.82 ($\pm$0.03) & 0.16 ($\pm$0.15) & 0.31 ($\pm$0.08)  & 0.38 ($\pm$0.05) & 0.0 ($\pm$0.0) \\
\bottomrule
\end{tabular}
\caption{Performance of our method, when substituting in different hindsight relabelling learning algorithms.}
\label{tab:igcbc}
\end{table}

In order to determine whether CRL is the appropriate goal-conditioned learning mechanism for the multi-agent setting, or whether this choice even matters beyond the use of hindsight relabelling, we train the following four baselines on the 2s3z SMAC environment (5 seeds): IHER, ICLearning (TD), ICL (Monte Carlo), and IGCBC. We preserve conventions from our formulation: sampling strategy, choice of $m_g$, discrete action parametrization, and network size. We find that these baselines perform significantly worse than ICRL on the task, as shown in the table above. 

We attribute this gap to ICRL’s representation learning formulation of Q function learning, which enables transfer across relabeled trajectories. In particular, ICL (MC) is identical to ICRL except that it uses a monolithic critic instead of two encoders. This method learns inconsistently with lower average performance than ICRL, supporting our hypothesis that temporal representations are essential to solve sparse-reward MARL tasks.

\include{checklist}

\end{document}